\documentclass[nohyperref]{article}

\usepackage{microtype}
\usepackage{graphicx}
\usepackage{subfigure}
\usepackage{booktabs} 

\usepackage{hyperref}

\usepackage[accepted]{icml2022}

\usepackage{amsmath}
\usepackage{amssymb}
\usepackage{mathtools}
\usepackage{amsthm}

\usepackage[capitalize,noabbrev]{cleveref}

\theoremstyle{plain}

\theoremstyle{definition}

\theoremstyle{remark}

\usepackage[textsize=tiny]{todonotes}

\usepackage{times}
\usepackage{soul}
\usepackage{url}
\usepackage[utf8]{inputenc}
\usepackage[small]{caption}
\usepackage{graphicx}
\usepackage{amsmath}
\usepackage{amsthm}
\usepackage{booktabs}
\usepackage{algorithm}
\usepackage{algorithmic}
\usepackage[utf8]{inputenc}
\usepackage{paralist}
\usepackage[shortlabels,inline]{enumitem}
\usepackage{graphicx}
\usepackage{xcolor}
\usepackage{hyperref}
\usepackage{multirow}

\usepackage[skip=1pt]{caption}

\newenvironment{mcsection}[1]
    {%
        \textbf{#1}

        \begin{itemize}[leftmargin=*,topsep=0pt,itemsep=-1ex,partopsep=1ex,parsep=1ex,after=\vspace{\medskipamount}]
    }
    {%
        \end{itemize}
    }

\usepackage{setspace}
\usepackage{changepage} 
\usepackage[breakable]{tcolorbox}
\usepackage{float} 

\newcommand{\pt}{Portal Telemedicina}

\newcommand{\header}[1]{\vspace{1mm}\noindent\textbf{#1}.}
\usepackage{balance}

\icmltitlerunning{Towards the Use of Saliency Maps for Explaining Low-Quality EKGs to End Users}

\begin{document}

\twocolumn[
\icmltitle{Towards the Use of Saliency Maps for Explaining \\ Low-Quality Electrocardiograms to End Users}



\icmlsetsymbol{equal}{*}

\begin{icmlauthorlist}
\icmlauthor{Ana Lucic}{pai,uva}
\icmlauthor{Sheeraz Ahmad}{goog}
\icmlauthor{Amanda Furtado Brinhosa}{pt}
\icmlauthor{Q. Vera Liao}{msr}
\icmlauthor{Himani Agrawal}{wiml}
\icmlauthor{Umang Bhatt}{moz,camb}
\icmlauthor{Krishnaram Kenthapadi}{fid}
\icmlauthor{Alice Xiang}{sony}
\icmlauthor{Maarten de Rijke}{uva}
\icmlauthor{Nicholas Drabowski}{pt}
\end{icmlauthorlist}

\begin{icmlauthorlist}
\icmlaffiliation{pai}{Partnership on AI, USA}
\icmlaffiliation{uva}{University of Amsterdam, Netherlands}
\icmlaffiliation{goog}{Google, USA}
\icmlaffiliation{pt}{Portal Telemedicina, Brazil}
\icmlaffiliation{msr}{Microsoft Research, Canada}
\icmlaffiliation{wiml}{WiMLDS, USA}
\icmlaffiliation{moz}{Mozilla Foundation, USA}
\icmlaffiliation{camb}{University of Cambridge, United Kingdom}
\icmlaffiliation{fid}{Fiddler AI, USA}
\icmlaffiliation{sony}{Sony AI, USA}
\end{icmlauthorlist}

\icmlcorrespondingauthor{Ana Lucic}{ana@partnershiponai.org}

\icmlkeywords{Machine Learning, ICML}

\vskip 0.3in
]



\printAffiliationsAndNotice{}  

\begin{abstract}
When using medical images for diagnosis, either by clinicians or artificial intelligence (AI) systems, it is important that the images are of high quality. 
When an image is of low quality, the medical exam that produced the image often needs to be redone. 
In telemedicine, a common problem is that the quality issue is only flagged once the patient has left the clinic, meaning they must return in order to have the exam redone.   
This can be especially difficult for people living in remote regions, who make up a substantial portion of the patients at \pt{}, a digital healthcare organization based in Brazil. 
In this paper, we report on ongoing work regarding
\begin{inparaenum}[(i)]
	\item the development of an AI system for flagging and explaining low-quality medical images in real-time, 
	\item an interview study to understand the explanation needs of stakeholders using the AI system at \pt{}, and
	\item a longitudinal user study design to examine the effect of including explanations on the workflow of the technicians in our clinics in the context of understanding low-quality medical exams. 
\end{inparaenum}
To the best of our knowledge, this would be the first longitudinal study on evaluating the effects of XAI methods on end-users -- stakeholders that use AI systems but do not have AI-specific expertise. 
We welcome feedback and suggestions on our experimental setup. 
\end{abstract}


\section{Introduction}
\label{section:introduction}

There exist many scenarios involving AI-assisted decision making in high-stakes industries such as healthcare \citep{elish_repairing_2020,van_leeuwen_artificial_2021,middleton_clinical_2016,litjens_survey_2017}. 
Explanations can help make such systems more transparent to various types of stakeholders \citep{mohseni2020multidisciplinary}. 
Prior work has found that there exists a significant gap between research and deployment for explainable AI (XAI), where current explanation techniques primarily cater to technical stakeholders rather than end users \citep{bhatt_explainable_2019}. 
In response, this work establishes a multistakeholder study with the goal of providing meaningful explanations to end users: individuals who interact with AI systems but do not have AI expertise themselves.

We first identify a real-world use case from \pt{},
a digital healthcare organization based in Brazil, where we believe explanations may be useful: flagging low-quality electrocardiogram (EKG) exams in real-time. 
Low-quality exams prevent clinicians from being able to accurately diagnose patients \citep{ahmed2011improving}, but are often not discovered until the end of the pipeline when they are forwarded to a clinician for diagnosis. 
At this point, many patients have already left the clinic, meaning they must return to the clinic if it turns out the exam needs to be redone. 
Given that many of our patients live in remote regions of Brazil, where it can be difficult to come to a clinic in the first place, it is important to be able to flag low-quality exams in real-time. 
We hypothesize that providing explanations along with the flags will help technicians understand the issues with the EKG exams so they can ensure a correct follow-up exam in a timely manner. 

In this paper, we adopt the 3-step approach recommended by \citet{bhatt_explainable_2019} for providing explanations to end users: 
\begin{inparaenum}[(i)]
    \item identifying stakeholders, 
    \item engaging with each stakeholder, and 
    \item understanding the purpose of the explanation.
\end{inparaenum}
We report on work in progress on developing, deploying and evaluating an AI system for flagging and explaining low-quality medical images. We describe the outcomes of two critical studies in our development process, aimed at answering the following research questions:
\begin{enumerate}[leftmargin=*,label=\textbf{RQ\arabic*:},nosep]
    \item What types of explanations are most appropriate for different types of stakeholders in the context of detecting low-quality medical exams?
    \item How can we evaluate explanations in (a) objective terms such as a user's ability to perform a task using an explanation, and (b) subjective terms such as the impact on a user's trust in an AI system?
\end{enumerate}

\noindent%
We answer \textbf{RQ1} by conducting an interview study with stakeholders from \pt{} to understand their explainability needs and goals. 
We use our qualitative analysis from the interviews to design a technician-facing interface for using our AI system for detecting and explaining low-quality medical images. 
That is, we use our answer from \textbf{RQ1} to design a system that we plan to test in \textbf{RQ2}. 
We answer \textbf{RQ2} by outlining the design and procedure for a large-scale, application-grounded \cite{doshi-2017-towards}, longitudinal study in order to evaluate the effect of including saliency map explanations on the workflow of technicians who perform EKG exams. 
We opt for a longitudinal study setup in order to be able to 
\begin{inparaenum}[(i)]
    \item evaluate the system as it would exist in the real world: integrated into their regular workflow, and
     \item evaluate if including such a system results in technicians performing better EKG exams over time. 
\end{inparaenum}
This is work in progress. We hope to obtain valuable feedback on our user study design through the workshop.


\section{Related Work}
\label{section:related-work}
Our work utilizes user studies for both the design of a medical (X)AI system (\textbf{RQ1}) and the design of an evaluation of a medical (X)AI system (\textbf{RQ2}). 
In the following subsections, we discuss prior work related to medical AI user studies (Section~\ref{section:medical-ai}) and medical XAI user studies (Section~\ref{section:medical-xai}). 

\subsection{Medical AI User Studies}
\label{section:medical-ai}

\textbf{Designing AI Systems}.
Recent years have witnessed a number of interview studies in the context of medical AI to elicitate the needs of professional end users and design medical AI systems based on their needs. 
For example, \citeauthor{lee2021_codesign}~[\citeyear{lee2021_codesign,lee2021_humanai}] design a human-AI collaborative system for stroke rehabilitation recommendation based on interviews with physical therapists who need to make such recommendations to their patients. 
\citet{Jacobs_2021_designing} design an AI decision support system for antidepressant treatment selection based on semi-structured interviews with physicians. 

Our work is similar to those mentioned above since it also designs an AI system for a medical task based on interviews with stakeholders involved in the development or use of the system. 
The main differences between these works and our work are:
\begin{inparaenum}[(i)]
    \item we focus on EKGs as medical images while previous works focus on other medical tasks, 
    \item our work includes an XAI component, and 
    \item our work also includes a proposal for a user study to evaluate the system.
\end{inparaenum}

\header{Designing and Evaluating AI Systems}
Other work with a similar setup to ours is by \citet{cai2019_imperfect}, who develop an AI system for retrieving similar medical images from previous patients in order to aid pathologists in diagnosis. 
Similar to our work, their work includes user studies for both the design and evaluation of the system. 
The main differences are that their work does not include an XAI component, or a longitudinal component.

\subsection{Medical XAI User Studies}
\label{section:medical-xai}

The use and effectiveness of explanations in medical AI is a topic of considerable recent interest. 
For example, \citet{tonekaboni2019clinicians} conduct an interview study with clinicians to understand their explainability needs and goals in intensive care units and emergency departments. 
In contrast, our work focuses on preventative medical care (i.e., medical screenings) as oppposed to acute medical care. 
Another distinction is that we use the findings from our interview study to implement an XAI system for end users, while the work of \citet{tonekaboni2019clinicians} is more exploratory in nature. 
We also propose a setup for evaluating our medical XAI system. 
Below, we detail recent work that utilizes user studies to design and evaluate medical XAI systems.

\header{Designing XAI Systems}
There have been several works which, similar to our work, develop XAI systems based on the needs of various types of stakeholders. 
\citet{cai_hello_2019} conduct an interview study to understand what information pathologists would like from an AI assistant when diagnosing prostate cancer as part of a human-AI collaborative decision making process. 
\citet{xie_chexplain_2020} develop an XAI system based on the needs of physicians and radiologists for exploring chest X-rays. 
In contrast, we focus on a different task: detecting low-quality EKGs. 

\citet{slack_2022_rethinking} interview doctors, healthcare professionals, and policymakers who already use AI explanations and find that these stakeholders prefer interactive explanations rather than static ones, specifically in the form of natural language dialogues. 
The authors subsequently propose a dialogue system for explainability in the medical domain. 
In contrast, we focus on static explanations because we are operating in a fairly low-resource setting and cannot accommodate the computational overhead of a sophisticated dialogue system.
Unlike the works mentioned above, we also propose a user study for evaluating the effects of our XAI system.

\header{Evaluating XAI Systems}
Although there have been many user studies in the fields of medical AI, medical XAI, and XAI more broadly, we are not aware of any other studies that investigate the effect of explanations through a longitudinal study. 
We note that our paper is a work in progress -- we propose a \emph{design} for a user study, while the works we list below report on the \emph{results} from their user studies.

\citet{pmlr-v126-hegselmann20a} investigate if generalized additive models, which should be ``inherently transparent'' from an AI point of view, can be safely interpreted by doctors. 
Similar to our work, they design a quantitative survey with end users (in their case, clinicians) to evaluate the effectiveness of their system.
This differs from our work in the medical task they focus on: predicting in-hospital mortality based on the first 48 hours of a patient's stay, as well as the absence of a longitudinal component.

\citet{taly_2019_using} evaluate saliency map explanations for diagnosing diabetic retinopathy with 10 ophthalmologists. 
\citet{jin2021_onemap} evaluate saliency map explanations for classifying brain tumours with 1 clinician. 
Our user study will also evaluate saliency map explanations, but our will be longitudinal and our user study will be on a larger scale.


\section{Problem Formulation}
\label{section:method}

\subsection{Task Description}

In this work, we focus on an AI system that helps end users (i.e., nursing technicians) identify low-quality EKG exams. 
Low-quality exams can arise due to a variety of factors such as mistakes on the user's part (e.g., putting electrodes in incorrect locations), technical issues (e.g., fraying wires), or patient errors (e.g., moving excessively during the exam). 
The AI system takes as input an image of the exam and outputs whether or not the exam is of low quality. 
The goal of the system is to flag low-quality exams in (near) real-time, so that the end user can redo the exam or take other remedial actions in a timely manner.

\begin{figure}[ht!]
    \centering
    \includegraphics[width=\columnwidth]{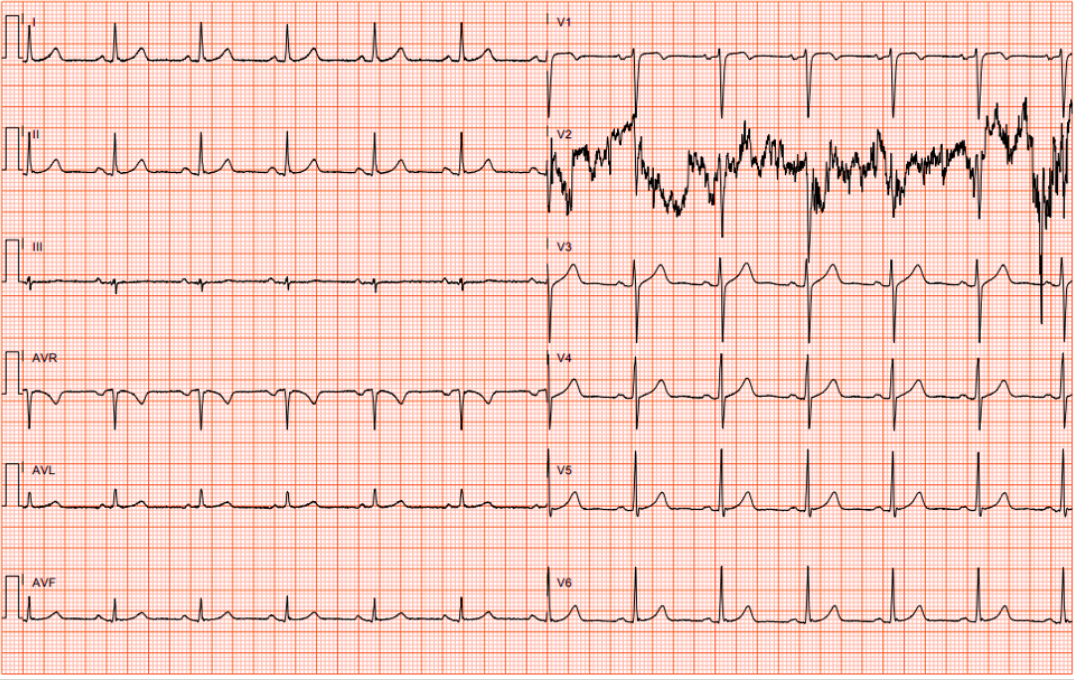}
    \includegraphics[width=\columnwidth]{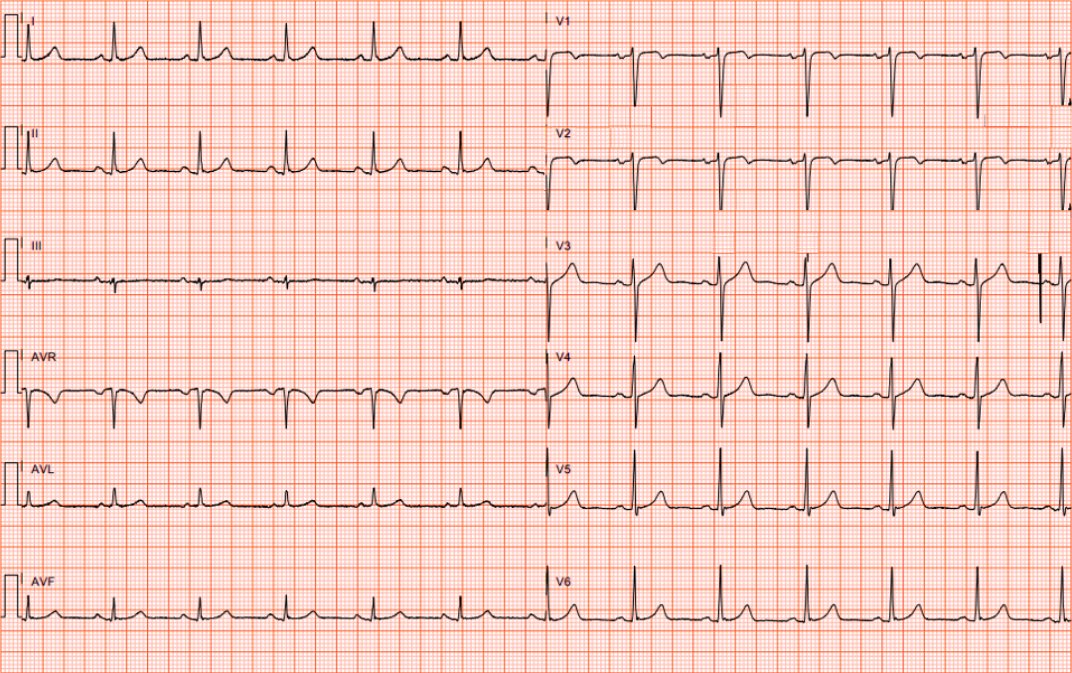}
     \includegraphics[width=\columnwidth]{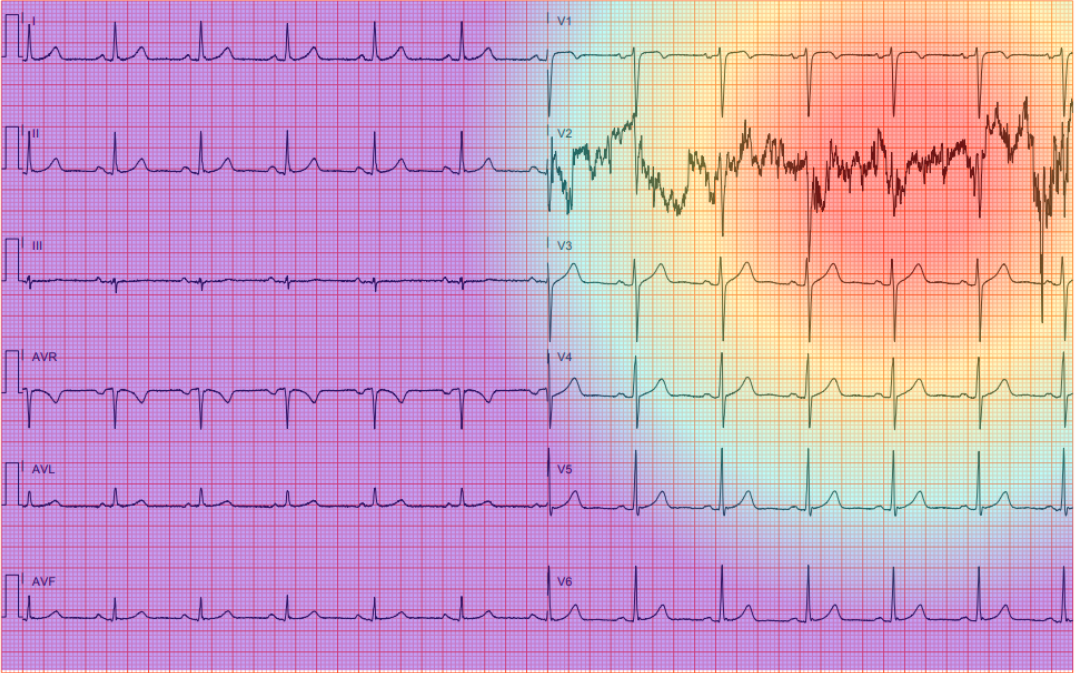}
    \caption{Behavior-based explanation probes for interview study. Top: Original example of a low-quality EKG scan. Middle: Counterfactual explanation. Bottom: Saliency map explanation for the original example.}
    \label{fig:explanation_examples}
\end{figure}

\subsection{Dataset and Model}
In general, the rate of low-quality medical exam across all of our clinics is approximately 7.5\%. 
Therefore, we first collect a balanced dataset from \pt{}'s proprietary database of historical EKG exams in order to train our ML model. The dataset consists of \emph{images} of EKG exams. 
We pull 10000 exams taken between 1 January 2020 to 8 September 2021, of which 5000 are low-quality. 
The binary low-quality label comes directly from the physicians who assess the exams: exams labelled as low-quality are unreadable by physicians. 
To avoid data leakage, we split the dataset into 80\% training, 10\% validation and 10\% test based on \textit{PatientID} (i.e., all exams from the same patient are in the same subset). 
Each patient has between 1 and 5 exams, with the vast majority (90\%) having only 1 exam.
The average age of patients in the dataset is 46 years old. 

To detect low-quality exams, we use transfer learning with the MobileNetV2 as the base: a convolutional neural network with inverted residual blocks and bottlenecking \cite{sandler_2018_mobilenet}. 
We train only the dense layers of MobileNetV2 using the Stochastic Gradient Descent (SGD) optimizer.
We use a fixed learning rate of 0.0001 and apply batch normalization after every layer. 
We train our model in batch sizes of 64 on one GPU, which takes approximately 2 days. 
Our model has 0.97 precision and 0.44 recall on the \emph{balanced} test set. 
This translates to 0.68 precision and 0.42 recall on the \emph{unbalanced} test set
For \pt{}, this is sufficient for the first version of our system.


\section{RQ1 Interview Study: Setup} 
\label{section:rq1-setup}

To answer \textbf{RQ1}, we conduct an interview study with different types of stakeholders from \pt{}.

\subsection{Study Design} 
We conducted 9 semi-structured interviews with participants who work at \pt{}: 2 executives, 3 developers, and 4 end users (i.e., technicians who perform medical exams). 
The group of participants had 3 women and 6 men. 
In order to understand the needs of various types of stakeholders involved in the process, the criteria for being included in the study was fairly broad: participants needed to have experience with an AI system, which could come in various forms such as development, deployment, interaction, or overseeing. 
Participants were recruited using internal communication tools at \pt{}. 
All participants completed a consent form before participating in the study and consented to being recorded during the interview.

\subsection{Procedure}
\label{section:rq1-setup}
The interviews were conducted online and lasted approximately 60 minutes. 
We provided the option of having a translator present during the interviews if the participants chose to do so. 
All questions were asked in English, which all the participants could understand, but some made use of the translator in order to express their responses in their native language. 
The interviews had six components:
\begin{inparaenum}[(i)]
    \item warm-up discussions, 
    \item understanding the task, 
    \item understanding end users, 
    \item user questions and requirements, 
    \item feedback on XAI features, and
    \item reflecting on XAI user needs. 
\end{inparaenum}
The full interview script is available in Appendix~\ref{app:A}.

\subsection{XAI Features}
The main purpose of the interview study is to understand the problem space and understand which explanations work best for which stakeholders. 
In the interview study, we ask participants to react to two types of explanations as defined by \citep{lucic2021multistakeholder}: 
\begin{inparaenum}[(i)]
	\item behavior-based, and 
	\item process-based. 
\end{inparaenum}
\emph{Behavior-based} explanations provide insight into how ML models make decisions (e.g., counterfactual examples \cite{wachter_counterfactual_2017,lucic_actionable_2020}, feature attributions \cite{ribeiro-2016-should,lundberg_unified_2017},  influential samples \cite{koh-2017-understanding,sharchilev-2018-finding}). 
\emph{Process-based} explanations provide insight into the whole ML modeling pipeline (e.g., model cards \cite{mitchell_model_2019}, datasheets \cite{gebru_datasheets_2020}). 


\begin{figure}[t!]
\begin{adjustwidth}{0pt}{0pt}
\begin{singlespace}

\tcbset{colback=white!10!white}

\begin{tcolorbox}[title=\textbf{Model Card: Low-Quality Exam Model},
     sharp corners, boxrule=0.7pt]

\begin{mcsection}{Model Details}
    \item Convolutional neural network based on MobileNetV2 \citep{sandler_2018_mobilenet}, implemented by \pt{} in 2021 for identifying low-quality EKG exams, input as images. 
\end{mcsection}

\begin{mcsection}{Intended Use}
    \item Model is intended for EKG scans from machines A and B, but not machine type C.
\end{mcsection}

\begin{mcsection}{Factors}
    \item Gender and age group. 
\end{mcsection}

\begin{mcsection}{Metrics}
    \item Accuracy, both over the whole population and within individual factors
\end{mcsection}

\begin{mcsection}{Training Data}
    \item Combination of data collected from a government database as well as scans taken at our clinics from years 2017--2018. 
    \item Preprocessing includes mean and standard normalization. 
\end{mcsection}

\begin{mcsection}{Evaluation Data}
    \item Same as training data, except from 2019--2020. 
\end{mcsection}

\begin{mcsection}{Ethical Considerations}
    \item Since human lives are involved, the Brazilian Health Regulatory Agency approved the development and research of this model. 
\end{mcsection}

\begin{mcsection}{Caveats and Recommendations}
    \item Although the model has high accuracy overall for people over 40, we do not have many data points for people 80+, so exercise caution when examining patients in this age group.
\end{mcsection}

\end{tcolorbox}
\end{singlespace}
\end{adjustwidth}
\caption{Process-based explanation probe for interview study: a mock model card for the low-quality exam model. }
\label{fig:model_card}
\end{figure}

Figure~\ref{fig:explanation_examples} shows the study probes we created for behavior-based explanations. 
We showed users an initial example of a low-quality EKG exam (Figure~\ref{fig:explanation_examples}: top). 
We then showed a counterfactual example (Figure~\ref{fig:explanation_examples}: middle), where the problematic part of the exam is replaced, in order to show a ``normal'' exam. 
Finally, we showed a feature attribution (i.e., a saliency map) that highlights the most important part of the original image in red and the least important parts of the image in purple, with a rainbow gradient in between (Figure~\ref{fig:explanation_examples}: bottom). We used a rainbow gradient because this aligns with what users from \pt{} have used in the past. 
Figure~\ref{fig:model_card} shows the mock model card we used as our process-based explanation probe in the interview study. 
In all cases, we showed the model card to users after showing the behavior-based explanation probes.


\section{RQ1 Interview Study: Results} 
\label{section:rq1-results}

The qualitative analysis of the interviews had three stages. 
First, several members of our team coded the same set of three interview transcripts (one for each type of stakeholder: executive, developer, and end user -- see Appendix~\ref{app:A} for details). 
Next, we consolidated a coherent set of themes, after which two members coded the rest of the interview transcripts according to the consolidated themes.

Table~\ref{table:themes} shows the eight main themes that emerged from the interview study outlined in Section~\ref{section:rq1-setup}. 
We group these themes into three broad categories: 
\begin{enumerate*}[label=(\roman*)]
    \item motivation,
    \item issues, and 
    \item desiderata.
\end{enumerate*} 
In the following subsections, we focus on some of the more prominent themes that came up during the interview study.

\subsection{Improving Outcomes}
Improving outcomes was seen by participants as one of the main motivations for including the low-quality flagging system in the pipeline. One participant broke this theme down into three main components:
\begin{quote}
\textit{``There are three benefits: benefit for the patient, because they don't need to go back to the clinic again, benefit for the clinic, there is, of course, the cost part of that, and improving the quality of the training of these technicians and the people that work with these exams.''}
\end{quote}

\noindent%
Given that many patients are coming into the clinics from remote regions, participants felt it was important to minimize the number of patients who need to return to the clinics due to low-quality exams:

\begin{quote}
\textit{``The idea is to use AI, not only for triage, but also to detect the technical problems fast enough so that we can send these results to the clinic before the patient leaves the clinic.''}
\end{quote}

\begin{table*}[h!]
\centering
\caption{Theme groupings from interview study. (\underline{Underlined themes} are discussed further in Section~\ref{section:rq1-results}.)}
\label{table:themes}
\begin{tabular}{ccc}
\toprule
Category 1: Motivation & Category 2: Issues  & Category 3: Desiderata \\ \midrule
\underline{Improving Outcomes}        & Challenges             & System Validity    \\
Perceived Benefits of XAI & Understanding Failures &    \underline{Trust in the System}   \\
Human-AI Cooperation      &                        &   \underline{Explanation Suitability}  \\
\bottomrule
\end{tabular}
\end{table*}

\subsection{Trust in the System}

The degree to which stakeholders trust the AI system for low-quality exam detection was another major theme that came up during the interview process. 
Almost all participants touched on some aspect of this theme, especially when it came to mistrusting the system, whether it was over-trusting or under-trusting:

\begin{quote}
\textit{``There is this pool of people that think that AI doesn't work, and they are not open for innovation. And there are other groups of people that think that the AI will provide better success, then they leave their work to the AI -- this is a problem too. We must bring both groups to the centre where they understand that the AI is trying to do a job but it's not perfect. It's an artificial \emph{intelligence}, so there is an artificial \emph`{dumbness'} associated too.'' }
\end{quote}

\begin{quote}
\textit{``With the doctors, we already saw that some of them think the AI will perform the work better than them, so they leave the work to the AI: this is a problem.''}
\end{quote}

\subsection{Explanation Suitability}
This theme emerged as a result of the questions we asked involving the interview probes shown in Figure~\ref{fig:explanation_examples} and Figure~\ref{fig:model_card}. 
Specifically, we wanted to understand which types of explanations were most useful for which stakeholders, which answers (\textbf{RQ1}). 
We found that the saliency maps (i.e., heatmaps) shown in Figure~\ref{fig:explanation_examples} (bottom) were the most favorably viewed explanations, across all types of stakeholders. 
All of the participants we interviewed found the saliency maps useful, and almost all of them believed that the saliency maps were the best option for technicians in the context of understanding why certain exams are predicted as being low-quality, including the technicians themselves. Therefore, we plan to test saliency maps explanations on our task of explaining low-quality EKG exams.  

\begin{quote}
\textit{``I think we should only show the heat maps: the less information, the better, and the smallest part of information we can deliver here are the heatmaps.''}
\end{quote}

\begin{quote}
\textit{``I think we should start only with heat maps. I think it's the simplest way to begin, technically.''}
\end{quote}

\begin{quote}
\textit{``Heatmaps are the most most friendly version of the explainability for the technicians.''}
\end{quote}

\noindent%
Our participants believed that counterfactual explanations, shown in Figure~\ref{fig:explanation_examples} (center), could be useful for understanding quality issues in EKG exams, especially when shown in combination with the saliency maps:
\begin{quote}
\textit{``It would be perfect to have them both to compare: the heat map and the counterfactual, because [the technicians] can see where the problem is with the heat map, and also an example of the correct exam with the counterfactual.''}
\end{quote}

\noindent%
As a result, we recommend using counterfactual explanations as a part of the training process for technicians, so they can learn to spot issues with low-quality exams by comparing them to exams that do not have quality issues. 
Some participants noted that counterfactual explanations could also be used by clinicians in an educational context to get a better understanding of how AI systems make decisions. 

The final type of explanation we tested was the mock model card shown in Figure~\ref{fig:model_card}. 
We found that most participants believed this type of explanation was best suited for stakeholders who need to have a more global view of the pipeline such as executives who make decisions about which models to productionize, or clinicians who use models to make diagnostic decisions about patients. 
None of our participants believed that model cards would be useful to technicians in the context of identifying low-quality exams in real-time, including the technicians themselves. 

\begin{quote}
\textit{``I think this [model cards], this solves some questions that doctors and healthcare professionals ask. They ask how many patients we used to train, how the data was collected, they ask all these questions. I personally think this would be good for them to have these answers.''}
\end{quote}

\noindent%
To sum up and answer \textbf{RQ1}: our participants believe that
\begin{enumerate*}[label=(\roman*)]
	\item saliency maps are useful for technicians who need to understand why certain exams are flagged as low-quality in real-time, 
	\item counterfactual explanations are useful as an educational tool -- either for technicians during their training, or for clinicians who are using an ML model to make patient-facing decisions, and
	\item model cards are useful for stakeholders who need to have a more global view of the modeling pipeline, such as executives or clinicians. 
\end{enumerate*}


\section{RQ2 Longitudinal Study: Setup} 
\label{section:rq2}

When examining \textbf{RQ1}, we found that stakeholders from \pt{} believed saliency map explanations could be useful for explaining low-quality EKG exams to end users. 
To answer \textbf{RQ2}, we outline the setup for a longitudinal, application-grounded \citep{doshi-2017-towards} study to examine the effect of saliency map explanations on the workflow of technicians. 

There are two components to our technician-facing system: 
\begin{inparaenum}[(i)]
    \item the low-quality prediction model, and
    \item the saliency map explanations. 
\end{inparaenum}
We plan to test three conditions:
\begin{itemize}
	\item Condition A: only model prediction 
	\item Condition B: model prediction + explanation
	\item Condition C: control (i.e., no input from AI system)
\end{itemize}
In our study, we will use saliency maps provided by GradCAM \citep{selvaraju2016grad} because they are straightforward to integrate into \pt{}'s pipeline. 
All technicians and clinics are located in Brazil and therefore this study was approved by the Brazilian Health Regulatory Agency.\footnote{\url{https://www.gov.br/anvisa/pt-br/english}}

\begin{figure*}
\centering
\includegraphics[scale=0.42]{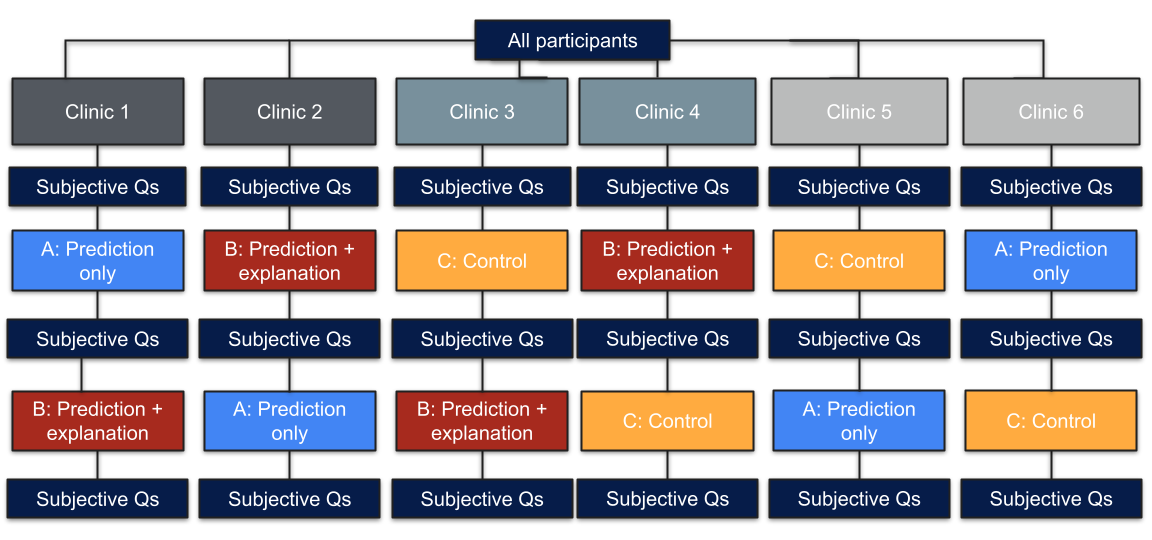}
\caption{Summary of longitudinal study setup: 6 clinics test 3 conditions (A, B, and C) in a block design.}
\label{fig:study_setup}
\end{figure*}

\subsection{Study Design} 
In this work, we opt for a longitudinal study design as opposed to a static study design in order to understand whether or not the system is worth integrating into the technicians' workflow. 
A static design would only provide us with information from a single snapshot in time, whereas we want to understand the effect of including such a system on the workflow of technicians \emph{over time}.

Evaluating our system is divided into two sub-goals: 
\begin{enumerate*}[label=(\roman*)]
	\item evaluating the technician's trust in the model prediction and its explanation, and
	\item evaluating how that translates to a lift in precision or recall of identifying low-quality exams. 
\end{enumerate*}

In order to evaluate (i), we quantify how often an exam needs to be redone following our system's prediction (and perhaps explanation) compared to the baseline of no interventions. This can depend on several subjective factors such as the perceived benefits of XAI, trust in the system, or other themes uncovered in our \textbf{RQ1} interview study (see Table~\ref{table:themes}).
A high agreement with the system is indicative of the technicians' trust. 

Similarly, in order to evaluate (ii), we compare precision, the ratio of correctly redone exams to all redone exams, as well as recall, the ratio of correctly performed redone exams to all low-quality exams -- across conditions A, B, and C. 
Improvement in precision signals better use of technician's time since fewer exams are redone unnecessarily. 
Improvement in recall signals better outcome for patients since the exams can be redone on the same day as opposed to a turn-around after a doctor's visit. 
We will obtain the ground truth labels directly from our clinicians in order to compute precision and recall. 
This is meant to answer \textbf{RQ2a}.

To answer \textbf{RQ2b}, a subjective questionnaire is given to gauge the understanding and level of comfort with using the new system. 
The questionnaire is based on the Likert-scale questions proposed in \cite{terhoeve-2017-news} and \cite{hoffman_metrics_2018}. 
See Appendix~\ref{app:B} for the full set of questions.

\subsection{Procedure}
\label{section:rq2-setup}

Since it is not feasible to assign individual technicians to treatment and control groups, we will instead assign treatment and control conditions to entire clinics. 
We cannot simply assign each clinic to one condition because our clinics vary in size, the number of patients that come in, and the number of technicians that work there. 
Therefore, each clinic will be subjected to two different conditions and we will switch the conditions halfway through the study. 
We will also need to control for the order in which the conditions are applied, meaning we need two clinics for each pair of conditions we want to test.  

Technicians are first given a brief introduction to machine learning, specifically how models can learn to perform classification and provide explanations in the form of saliency maps. 
They are also given hands-on training on how to access the new interface for accessing model's prediction and explanations, and on reaching tech support when needed. 
During the study, the technicians will continuously interact with the system as part of their day-to-day jobs. 

For each patient, the technicians will perform the EKG exam. 
If the technician's clinic is under one of the treatment conditions (A or B) and the model predicts the exam is low-quality, then the technician has to make two decisions that are logged explicitly through a button in the interface:
\begin{itemize}
	\item Do they agree with the model or override the model?
	\item Do they redo the exam or leave the original exam?
\end{itemize}
Although the answers to these two questions would usually align (agreeing with the model implies redoing the exam), there are some emergency situations where they may not, which is why we log them separately. 

The \textbf{RQ2b} subjective questions will be administered at three touch points: 
\begin{enumerate*}[label=(\roman*)]
	\item at the beginning of the study, 
	\item after the conditions switch, and 
	\item at the end of the study. 
\end{enumerate*}

\section{Conclusion}
\label{section:conclusion}

In this work, we have reported on work in progress regarding our AI system for detecting and explaining low-quality EKG exams at \pt{}. 
We first identify which types of explanations are most appropriate for our use case by conducting a user study with stakeholders from \pt{}, in order to understand their explainability needs and goals. 
Next, we outline the setup for an application-grounded, longitudinal study with end users from \pt{} in order to evaluate our system for AI-based detection of low-quality scans, 
For future work, we will test the effectiveness of including saliency map explanations on the workflow of technicians in our clinics and hopefully improve diagnostic outcomes for our patients. 

\pagebreak

\section*{Acknowledgements}
This research was supported by the Partnership on AI, the Netherlands Organisation for Scientific Research (NWO) under project nr.\ 652.\-001.\-003, DeepMind and the Leverhulme Trust via the Leverhulme Centre for the Future of Intelligence (CFI), the Mozilla Foundation, and the Hybrid Intelligence Center, a 10-year program funded by the Dutch Ministry of Education, Culture and Science through the NWO, \url{https://hybrid-intelligence-centre.nl}. %
All content represents the opinion of the authors, which is not necessarily shared or endorsed by their respective employers and/or sponsors.

\bibliography{xai_bib}
\balance
\bibliographystyle{icml2022}

\newpage
\appendix
\onecolumn
\section*{Appendix}

\section{RQ1 Interview Study Script}
\label{app:A}
Below is the full interview script for the \textbf{RQ1} study. Some groups of questions were more applicable to certain types of stakeholders than others, which we indicate in parentheses. 

\begin{enumerate}[leftmargin=*]
    \item Warm-up discussions:
    \begin{itemize}
        \item Can you describe your role, how long you've been with the company, and what you're working on?
        \item Can you describe what the low-quality scan model is?
    \end{itemize}
    \item Understanding the task (primarily for developers): 
    \begin{itemize}
        \item What part of the the low-quality scan model have you worked on?
        \item Can you explain what type of model it is? What data is it trained on?
        \item What is considered a low-quality scan? How often does it happen that a scan is not good enough and needs to be redone? What are the common reasons for this?
        \item How do you usually identify a low-quality scan? What happens if it is not identified on the spot?
    \end{itemize}
    \item Understanding end users (primarily for developers and executives): 
    \begin{itemize}
        \item Can you describe who the target users are?
        \item Have you interacted with the end users directly, or learned about them?
        \item What do you believe is the main value that the low-quality scan model would deliver, or the main user problem it solves?
        \item Do you foresee any challenges for the users to use the low-quality scan model?
        \item What factors do you think might determine whether users would adopt or trust the low-quality scan model? Is the product team doing anything to enhance user adoption or trust?
        \item Besides what we discussed, are there any other user problems or design issues the team is prioritizing to solve for the low-quality scan model?
        \item What does explainability mean to you in the context of the low-quality scan model? Why does your team consider it a priority?
    \end{itemize}
    \item User questions and requirements: 
    \begin{itemize}
        \item Imagine you are a nurse or technician working with the low-quality scan model, what kind of questions would you ask of the system?
        \item Why do you think users would want to ask that? What would a good answer look like? What would a good answer achieve?
        \item Are there any other questions that the system should be able to answer in order for users to use it and trust it?
    \end{itemize}
    \item XAI features:
    \begin{itemize}
        \item What are some examples of XAI features that the product team has considered, or are currently developing? For each one, we ask:
        \begin{itemize}
            \item When do you think users might need this XAI feature? How can it help the users?
	        \item How did the team come up with this XAI feature?
	        \item Do you foresee any challenges or problems with this kind of XAI feature?
        \end{itemize}
        \item We show users 3 examples of XAI features: saliency maps, counterfactual examples and model cards (see Figure~\ref{fig:explanation_examples} and Figure~\ref{fig:model_card}). For each XAI feature, we ask: 
        \begin{itemize}
            \item Do you think users might need this XAI feature? Would it help the users?
            \item Do you foresee any challenges or problems with this kind of XAI feature?
        \end{itemize}
    \end{itemize}
    \item Reflecting on XAI features and user needs:
    \begin{itemize}
        \item Are these XAI features enough, or do you foresee any challenges that we have not covered?
    \item Are there any other XAI features or information that you think the low-quality scan model could provide?
    \item For developers only: What kinds of XAI features would you find useful for developing or debugging the models? Have you used any? What was your experience?
    \end{itemize}
\end{enumerate}

\section{RQ2b Subjective Questions}
\label{app:B}

Below is the full list of subjective questions for \textbf{RQ2b}. 
\begin{itemize}
	\item I understand why the prediction is low-quality. 
	\item I support using this system as a tool. 
	\item I trust this system. 
	\item In my opinion, this system produces mostly reasonable outputs. 
	\item I am confident in the system. I feel that it works well.
	\item The outputs of the system are very predictable.
	\item The system is very reliable. I can count on it to be correct all the time.
	\item I feel that when I rely on the system, I will get the right answers.
	\item I am wary of the system.
	\item I like using the system for decision making.
\end{itemize}



\end{document}